\documentclass{article}
\usepackage{ijcai18}

\usepackage{times}
\usepackage{xcolor}
\usepackage{soul}
\usepackage[utf8]{inputenc}
\usepackage[small]{caption}

\usepackage{hyperref}
\usepackage{url}
\usepackage{subfigure}
\usepackage{graphicx}  
\usepackage{multicol}
\usepackage{multirow}
\usepackage{amsmath,amssymb}
\newcommand{\tabincell}[2]{\begin{tabular}{@{}#1@{}}#2\end{tabular}}
\usepackage{slashbox}
\usepackage{epstopdf}

\title{Cross-domain Dialogue Policy Transfer via \\Simultaneous Speech-act and Slot Alignment}

\author{
Kaixiang Mo$^1$,
Yu Zhang$^2$,
Qiang Yang$^3$,
Pascale Fung$^4$,
\\
Hong Kong University of Science and Technology, Hong Kong, China \\
\{kxmo, yuzhangcse, qyang\}@cse.ust.hk,
pascale@ece.ust.hk
}

\begin{document}

\maketitle

\begin{abstract}
Dialogue policy transfer enables us to build dialogue policies in a target domain with little data by leveraging knowledge from a source domain with plenty of data. Dialogue sentences are usually represented by speech-acts and domain slots, and the dialogue policy transfer is usually achieved by assigning a slot mapping matrix based on human heuristics.
However, existing dialogue policy transfer methods cannot transfer across dialogue domains with different speech-acts, for example, between systems built by different companies. Also, they depend on either common slots or slot entropy, which are not available when the source and target slots are totally disjoint and no database is available to calculate the slot entropy.
To solve this problem, we propose a Policy tRansfer across dOMaIns and SpEech-acts (PROMISE) model, which is able to transfer dialogue policies across domains with different speech-acts and disjoint slots. The PROMISE model can learn to align different speech-acts and slots simultaneously, and it does not require common slots or the calculation of the slot entropy.
Experiments on both real-world dialogue data and simulations demonstrate that PROMISE model can effectively transfer dialogue policies across domains with different speech-acts and disjoint slots.
\end{abstract}

\section{Introduction}
Based on different models used, task-oriented dialogue systems can be divided into rule-based and learning-based systems.
Learning-based task-oriented dialogue systems~\cite{young2013pomdp,williams2016end,wen2016network,li2016deep,serban2015hierarchical,mou2016sequence}, which is the focus of our research, can learn robust dialogue policies from training data without handcrafted dialogue decisions made by human. One requirement of learning-based task-oriented dialogue systems is the availability of a lot of training dialogues which are used to train the dialogue policy, however, in many applications, this requirement is hard to satisfy as labeling data is laborious and time costly. In this situation, a good solution is to use transfer learning \cite{pan2010survey} to alleviate the data sparsity problem.

Transfer learning can be used to build a dialogue policy on a target domain with limited data by leveraging dialogue policy and dialogue data from a source domain.
Specifically, the sentences in task-oriented dialogue systems are usually represented by speech-acts~\cite{DBLP:journals/ai/AllenP80} and domain slots. For example, ``I want to book a four-star hotel'' can be represented by ``inform(type=hotel, stars=4star)'', where ``inform'' is a speech-act, ``type'' and ``stars'' are slots related to the hotel booking domain, and ``hotel'' and ``4star'' are specific slot values for the two slots.
Many dialogue policy transfer methods have been proposed.
Ga{\v{s}}i{\'c} \textit{et al.}~\cite{gavsic2013pomdp,gavsic2014incremental} propose to adapt a dialogue policy to an extended domain with additional slots with policy fine-tuning and using the source domain dialogue policy as a prior. In this case, the source domain and the target domain should have a considerable portion of common slots, or the similarity values between new slots and old slots have to be manually assigned.
Ga{\v{s}}i{\'c} \textit{et al.}~\cite{gavsic2015policy,gavsic2015multi} propose to build cross-domain slot similarity matrix by using a normalized entropy. 
The calculation of the normalized entropy for a slot requires to access all entities in a database. However, in many situations such database is usually not available. Moreover, existing dialogue policy transfer learning methods cannot transfer across dialogue systems with a different set of speech-acts.

However, existing dialogue policy transfer methods cannot transfer across dialogue domains with different speech-acts. For example, the speech-act ``inform'' in a dialogue system might correspond to the speech-act ``tell'' or even ``Action1'' in another dialogue system. Moreover, existing dialogue transfer methods do not work when the source and target domains do not have common slots or when no database can be used to calculate the normalized entropy between slots.

We propose a Policy tRansfer across dOMaIns and SpEech-acts (PROMISE) model, which is able to transfer dialogue policies across domains with different speech-acts and disjoint slots. Specifically, the proposed PROMISE model learns a cross-domain speech-act similarity matrix and a slot similarity matrix by optimizing the performance of the transferred Q-function on the target domain data, so that the model can learn the cross-domain mapping with optimal target domain performance.
Different from existing works, the PROMISE model does not require common slots or a database to calculate the normalized entropy for pairs of slots. Extensive simulated and real-world experiments show that the PROMISE model can effectively transfer dialogue policies.

The contributions of this paper are three-fold:
\begin{enumerate}
\item We define and formulate the simultaneously cross-speech-act and cross-domain dialogue policy transfer problem.
\item We propose a novel transfer learning model, the PROMISE, for the problem. The PROMISE model does not require common slots, common speech-acts or full database access to calculate normalized entropy for each slot. Instead it learns a cross-domain mapping that can maximize the performance of the transferred policy from a few target-domain data.
\item We conduct simulation experiments and collect real-world cross-domain dialogue datasets to evaluate the proposed algorithm and the experimental results show that the proposed model can effectively transfer dialogue policies across domains.
\end{enumerate}

\section{Related Works}
\label{crossTL:Related Work}

Transfer learning has been used in dialogue system to solve the data sparsity problem in spoken language understanding, dialogue state tracking, dialogue policy learning~\cite{gavsic2013pomdp,gavsic2014incremental,gavsic2015policy,gavsic2015multi} and natural language generation. In this paper, we focus on dialogue policy transfer learning problems on multi-turn task-oriented dialogue systems.
Transfer learning techniques for reinforcement learning~\cite{taylor2009transfer} have been used in various domains, but most methods require a predefined state mapping or action mapping assigned by human experts.
Taylor \textit{et al.}~\cite{taylor2008autonomous} propose to test all possible state and action mapping offline and choose the mapping that can minimize the prediction error of the target model on the source data. However, the number of all possible state and action mapping grows exponentially in terms of the number of speech-acts and slots, so this algorithm is computationally inefficient and impractical for real-world task-oriented dialogue systems. Mo \textit{et al.}~\cite{mo2016personalizing,mo2017fine} propose to transfer dialogue policies across different users by modelling a dialogue policy with a common and a personal part, but these models cannot transfer across different domains with different slots and speech-acts.

In contrast to the above mentioned works, the proposed PROMISE model is able to transfer dialogue policies across domains with different slots and speech-acts and it does not require the source and target domains to share common slots or an additional database to calculate the entropy. 

\section{Problem Settings}
\label{crossTL:Problem}

In this paper, matrices are denoted in the bold capital case, vectors are in the bold lower case, and scalars are in the lower case. The utterance of the user is denoted by $X$, and the utterance of the agent is denoted by $Y$. A multi-turn dialogue data is represented as $\{X_n, Y_n, r_n\}$, where $X_n$ and $Y_n$ are the user utterance and the agent reply in the $n$-th dialogue turn and $r_n$ is the immediate reward for the $n$-th dialogue turn.

Given a target domain with a few training dialogues and a source domain with plenty of training dialogues, the goal is to learn a dialogue policy in the target domain by leveraging knowledge in the source domain.
The inputs for this problem include:
\begin{enumerate}
\item Plenty of source domain training dialogues $\mathcal{D}^s=\{X^s_n, Y^s_n, r_n\}$.
\item A few target domain training dialogues $\mathcal{D}^t=\{X^t_n, Y^t_n, r_n\}$.
\end{enumerate}
The output of the problem is:
\begin{enumerate}
\item A dialogue policy $\pi^t$ for the target domain.
\end{enumerate}
Note that in this problem, the speech-acts and domain slots in the source and target domains can be totally different and no external database is available.

\section{PROMISE Model}
\label{crossTL:Model}
In this section, we will first introduce the single-domain dialogue system and then introduce the proposed PROMISE model. We build the PROMISE model based on the PyDial package~\cite{ultes2017pydial}.

\begin{table*}[ht]
\centering
\caption{Different abstractions for sentences and dialogue states.}
\label{crossTL:tab:abstract}
\begin{tabular}{|l|l|l|l|l|}
\hline
\tabincell{l}{Abstraction} & Notation & Description & Example  \\ \hline
\tabincell{l}{Original} & $X_n,Y_n,H_n$ & \tabincell{l}{$H_{n}=\{\{X_k, Y_k\}_{k=1}^{n-1}, X_n\}$} & \tabincell{l}{$X_1$=``I want to find an expensive hotel''}  \\ \hline
\tabincell{l}{Abstracted} & $\tilde{X}_n,\tilde{Y}_n, \tilde{H}_n$ & \tabincell{l}{$\tilde{X}_n=\{a_n, \{s_k=v_k\}_{k=1}\},$ \\$\tilde{Y}_n=\{\bar{a}_n, \{\bar{s}_k=\bar{v}_k\}_{k=1}\},$ \\$\tilde{H}_n=\{l_{n}, a_n, \{\tilde{s}_k,\tilde{v}_k\}, \{\hat{s}_k,\hat{v}_k\}\}$} & \tabincell{l}{$\tilde{X}_1$=``inform(type=hotel, price=expensive)'';\\$\tilde{H}_1=\{l_{n}=10, a_1=\text{inform},$\\$ \{\tilde{s}_1=\text{price}, \tilde{v}_1=\text{expensive}\}, \{\}\}$}  \\ \hline
\tabincell{l}{Summary} & $\mathbf{x}_n,\mathbf{y}_n,\mathbf{h}_n$ & \tabincell{l}{$\mathbf{x}_n=[\mathbf{a}_n, \mathbf{s}_n],$ $\mathbf{y}_n=[\bar{\mathbf{a}}_n, \bar{\mathbf{s}}_n],$ \\$\mathbf{h}_n=[\mathbf{l}_{n}, \mathbf{a}_n, \tilde{\mathbf{s}}_n, \hat{\mathbf{s}}_n]$} & \tabincell{l}{$\mathbf{x}_1=[[0,0,0,1],[0,1,0,0]];$\\$\mathbf{h}_1=[[10], [0,0,0,1], [0,1,0,0], [0,0,0,0]]$}  \\ \hline
\end{tabular}
\end{table*}

\subsection{Abstractions in Sentences and Dialogue States}
In order to make models have better generalization ability, there are three levels of abstraction in sentences and dialogue states in the PyDial~\cite{ultes2017pydial} package.

An \textbf{original} sentence is represented by a sequence of \mbox{words}. The original dialogue state for the $n$-th agent reply $Y_n$ is defined as $H_n=\{\{X_k, Y_k\}_{k=1}^{n-1}, X_n\}$, which is the collection of all historical user utterances and agent replies in the current dialogue session.

An \textbf{abstracted} sentence is represented by a speech-act $a$, a collection of slots $\{s\}$ and their slot values $\{v\}$. An abstracted user utterance is denoted by $\tilde{X}_n=\{a_n, \{s_k=v_k\}\}$ and an abstracted agent reply is denoted by $\tilde{Y}_n=\{\bar{a}_n, \{\bar{s}_k=\bar{v}_k\}\}$.
The abstracted dialogue state for the $n$-th agent reply is denoted by $\tilde{H}_n=\{l_{n}, a_n, \{\tilde{s}_k,\tilde{v}_k\}, \{\hat{s}_k,\hat{v}_k\}\}$, where $l_{n}$ is the number of entities that match all user constraints, $a_n$ is the user speech-act in the $n$-th turn, $\{\tilde{s}_k,\tilde{v}_k\}$ is the collection of user constraints and $\{\hat{s}_k,\hat{v}_k\}$ is the collection of user requests. For example, ``I want an expensive hotel'' is a user constraint and ``what is the address'' is a user request.

A \textbf{summary} sentence~\cite{thomson2010bayesian,young2010hidden} in the dialogue policy module is represented by a speech-act one-hot vector $\mathbf{a}_n$ and a slot one-hot vector $\mathbf{s}_n$ without slot values. 
A summary user utterance is denoted by $\mathbf{x}_n=[\mathbf{a}_n, \mathbf{s}_n]$ and a summary agent reply is denoted by $\mathbf{y}_n=[\bar{\mathbf{a}}_n, \bar{\mathbf{s}}_n]$.
The summary dialogue state for the $n$-th agent reply is denoted by $\mathbf{h}_n=[\mathbf{l}_{n}, \mathbf{a}_n, \tilde{\mathbf{s}}_n, \hat{\mathbf{s}}_n]$, where $\mathbf{l}_{n}$ is a one-hot state vector indicating the number of entities matching current user's constraints, $\mathbf{a}_n$ is the user speech-act vector in the $n$-th turn, $\tilde{\mathbf{s}}_n$ is the user constraint vector indicating whether the user has provided information about each slot so far, and $\hat{\mathbf{s}}_n$ is a vector indicating whether the user has request information about each slot. The details for the three abstractions are summarized in Table~\ref{crossTL:tab:abstract}.

\subsection{Single-Domain Dialogue System}
The spoken language understanding (SLU) module transforms user utterance $X_n$ into the abstracted user utterance $\tilde{X}_n$, i.e., $\tilde{X}_n = \text{SLU}(X_n)$, and it is based on the regular expression for efficiency. The dialogue state tracking (DST) module updates current dialogue state $\tilde{H}_n$ based on the previous dialogue state $\tilde{H}_{n-1}$, the previous abstracted system utterance $\tilde{Y}_{n-1}$ and the abstracted user utterances $\tilde{X}_{n}$, i.e., $\tilde{H}_n = \text{DST}(\tilde{H}_{n-1}, \tilde{Y}_{n-1}, \tilde{X}_{n})$.
The dialogue policy model (DPL) can choose the agent reply $\tilde{Y}_n$ to maximize the total expected reward, i.e., $\tilde{Y}_n = \pi(\tilde{H}_n)$.
The natural language generation (NLG) module converts the abstracted agent reply in speech-act slot representation $\tilde{Y}_n$ into a fluent sentence $Y_n$, i.e., $Y_n = \text{NLG}(\tilde{Y}_n)$.

\subsection{The PROMISE Model}
In this section, we introduce the PROMISE model. The PROMISE model can learn to map dialogue states and candidate agent replies from the target domain to the source domain, and then leverage the Q-function in the source domain.

The Q-function of the PROMISE model can be formulated as
\begin{equation*}
Q^t(\mathbf{h}^t, \mathbf{y}^t) = Q^s(f^{t2s}_h(\mathbf{h}^t), f^{t2s}_y(\mathbf{y}^t)) ,
\end{equation*}
where $Q^s(\mathbf{h}^s, \mathbf{y}^s)$ is the source domain dialogue policy, $f^{t2s}_h(\mathbf{h}^t)$ is the cross-domain state translation function, and $f^{t2s}_y(\mathbf{y}^t)$ is the cross-domain sentence translation function. The source domain dialogue policy can be any dialogue policy and in this paper we utilize the Gaussian process dialogue policy~\cite{gavsic2014gaussian}.

Before defining the cross-domain state translation function $f^{t2s}_h(\mathbf{h}^t)$ and the cross-domain sentence translation function $f^{t2s}_y(\mathbf{y}^t)$, we need to firstly define the cross-domain speech-act mapping function $f^{t2s}_a(\mathbf{a}^t)$ and the cross-domain slot mapping function $f^{t2s}_s(\mathbf{s}^t)$.


Given a speech-act vector $\mathbf{a}^t$ in the target domain, the translated source speech-act vector $\mathbf{a}^s$ is
\begin{equation*}
\mathbf{a}^s = f^{t2s}_a(\mathbf{a}^t) =  \mathbf{a}^t \mathbf{M}_a^{t2s},
\end{equation*}
where $\mathbf{M}_a^{t2s}$ is the speech-act similarity matrix from the target domain to the source domain and its entry $\bar{m}_{ij}^{t2s}$ in the $i$-th row and $j$-th column is the speech-act similarity between the $i$-th target speech-act and $j$-th source speech-act. Note that $\mathbf{a}^s$ and $\mathbf{a}^t$ are treated as probability vectors where the sum of all entries in $\mathbf{a}^s$ and $\mathbf{a}^t$ equals $1$. 
Hence $\bar{m}_{ij}^{t2s}$ can be parametrized as $\bar{m}_{ij}^{t2s} = \exp(\mathbf{e}_{a^t_i}\mathbf{e}_{a^s_j}^T) / {\sum_{\forall j}\exp(\mathbf{e}_{a^t_i}\mathbf{e}_{a^s_j}^T})$, where $\mathbf{e}_{a^t_i}$ is the embedding vector of speech-act $a^t_i$ and the superscript $^T$ denotes the matrix transpose. $f^{s2t}_a(\cdot)$ can be defined similarly.

Given a slot vector in the target domain $\mathbf{s}^t$, the translated source slot vector $\mathbf{s}^s$ is
\begin{equation*}
\mathbf{s}^s = f^{t2s}_s(\mathbf{s}^t) = \mathbf{s}^t \mathbf{M}_s^{t2s},
\end{equation*}
where $\mathbf{M}_s^{t2s}$ is the slot similarity matrix from the target domain to the source domain and its entry $m_{ij}^{t2s}$ in the $i$-th row and $j$-th column is the slot similarity between the $i$-th target slot and the $j$-th source slot. Note that $\mathbf{s}^s$ and $\mathbf{s}^t$ can be viewed probability vectors as all entries in $\mathbf{s}^s$ and $\mathbf{s}^t$ sum to $1$. 
$m_{ij}^{t2s}$ is parametrized as $m_{ij}^{t2s} = \exp(\mathbf{e}_{s^t_i}\mathbf{e}_{s^s_j}^T) / {\sum_{\forall j}\exp(\mathbf{e}_{s^t_i}\mathbf{e}_{s^s_j}^T})$, where $\mathbf{e}_{s^t_i}$ is the embedding vector of slot $s^t_i$ and the subscript $T$ is matrix transpose. $f^{s2t}_s(\cdot)$ can be defined similarly.

The cross-domain summary sentence translation function $f^{t2s}_y(\cdot)$ from the target sentence $\mathbf{y}^t = [\bar{\mathbf{a}}^t, \bar{\mathbf{s}}^t]$ to the corresponding source sentence $\mathbf{y}^s = [\bar{\mathbf{a}}^s, \bar{\mathbf{s}}^s]$ can be defined as
\begin{align*}
\mathbf{y}^s = f^{t2s}_y(\mathbf{y}^t) = [f^{t2s}_a(\bar{\mathbf{a}}^t), f^{t2s}_s(\bar{\mathbf{s}}^t)],
\end{align*}
and $f^{s2t}_y(\cdot)$ can be defined similarly.

The cross-domain state mapping function $f^{t2s}_h(\mathbf{h}^t)$ from the target state $\mathbf{h}^t = [\mathbf{l}, \mathbf{a}^t, \tilde{\mathbf{s}}^t, \hat{\mathbf{s}}^t]$ to the corresponding source state $\mathbf{h}^s = [\mathbf{l}, \mathbf{a}^s, \tilde{\mathbf{s}}^s, \hat{\mathbf{s}}^s]$ can be defined as
\begin{align*}
\mathbf{h}^s = f^{t2s}_h(\mathbf{h}^t) = [\mathbf{l}, f^{t2s}_a(\mathbf{a}^t), f^{t2s}_s(\tilde{\mathbf{s}}^t), f^{t2s}_s(\hat{\mathbf{s}}^t)] ,
\end{align*}
where $\mathbf{l}$ in both the source and the target domain have the same format. $f^{s2t}_h(\cdot)$ can be defined similarly.

\subsection{Parameter Learning}
In this section, we introduce the objective function used in the PROMISE model. We optimize the objective function with the Adam algorithm~\cite{kingma2014adam}.

\subsubsection{Reward Function}
If the user finds a entity that matches his constraints or is told that there is no such entity within $20$ turns, the dialogue is successful and the agent will get a final reward of $20$. 
For each dialogue turn, the agent receives an immediate reward of $-1$ to punish longer dialogues. This reward function is offered by the Pydial package~\cite{ultes2017pydial} and the PROMISE model can be used with other reward functions as well. 

\subsubsection{Loss Function}
We combine the Q-learning~\cite{watkins1989learning} objective with a series of regularization terms as the final loss function, which is defined as
\begin{equation*}
L(\Theta) = \mathbb{E}[r_n + \gamma \max_{\mathbf{y}'} Q^t(\mathbf{h}_{n+1}, \mathbf{y}') - Q^t(\mathbf{h}_{n}, \mathbf{y}_n) ]^2 + \mathcal{R}(\Theta) ,
\end{equation*}
where $\Theta$ denotes the set of parameters and $\mathcal{R}(\Theta)$ is the regularization term which will be introduced in the next section.

\subsubsection{Regularizations}
$\mathcal{R}(\Theta)$ is the sum of four regularizations terms used for learning the cross-domain translation functions.

Note that each speech-act $\mathbf{a}^s$ co-occurs with only some of the slots and hence we can predict the slot based on the speech-act. Specifically, firstly we train a prediction function $c^t(\cdot)$ in the target domain to predict the slot vector $\mathbf{s}^t$ conditioned on the speech-act vector $\mathbf{a}^t$, then the cross-domain slot vector preservation regularizer can be defined as
\begin{equation*}
\mathcal{R}_{1s}(\Theta) = \frac{1}{|\mathcal{D}^s|} \sum_{\{\mathbf{a}^s,\mathbf{s}^s\} \in \mathcal{D}^s} L_{ce}( (f^{t2s}_s \circ c^t \circ f^{s2t}_a)(\mathbf{a}^s), \mathbf{s}^s)
\end{equation*}
where $L_{ce}(\cdot)$ denotes the cross-entropy loss and $\circ$ denotes the functional composition. In this regularizer, $f^{s2t}_a (\mathbf{a}^s)$ can be viewed as the corresponding similar speech-act vector in the target domain to $\mathbf{a}^s$ and $(c^t \circ f^{s2t}_a)(\mathbf{a}^s)$ is the predicted compatible slot vector in the target domain. Similarly, we can define $\mathcal{R}_{1t}(\Theta)$ in the target domain. The cross-domain slot vector preservation regularizer can be defined as $\mathcal{R}_1(\Theta) = \mathcal{R}_{1s}(\Theta) + \mathcal{R}_{1t}(\Theta)$.

Since a user would reply according to the question being asked, we can train a prediction function $c^s_u(\cdot)$ in the source domain to predict the user's next speech-act $\mathbf{a}_{n}$ based on the system reply $\mathbf{y}_{n-1}$. Then the cross-domain user prediction regularizer can be expressed as
\begin{equation*}
\mathcal{R}_2(\Theta) = \frac{1}{|\mathcal{D}^t|} \sum_{\{\mathbf{a}_n,\mathbf{y}_{n-1}\} \in \mathcal{D}^t} L_{ce}( (f^{s2t}_a \circ c^s_u \circ f^{t2s}_y)(\mathbf{y}_{n-1}), \mathbf{a}_{n}).
\end{equation*}

Since a speech-act should have similar occurrence probabilities after cross-domain translation, the cross-domain frequency regularizer of both user speech-act and the \mbox{agent} speech-act is defined as
\begin{equation*}
\mathcal{R}_3(\Theta) = L_{kl}(f^{t2s}_a(\mathbf{p}_a^t), \mathbf{p}_a^s) ,
\end{equation*}
where $\mathbf{p}_a^t$ is the occurrence probability of all the target domain speech-acts, $\mathbf{p}_a^s$ is the occurrence probability of all the source domain speech-acts and $L_{kl}()$ is the Kullback–Leibler divergence loss. 

In a one-to-one slot mapping, some slots cannot be mapped to any slot when the target and source domains have different numbers of slots, so more than one target dialogue state might correspond to the same source domain state.
In order to allow updates of all slots in the target state-vector to affect the corresponding translated source domain state-vector, we encourage each target slot to map to as many source slots as possible and hence the state continuity regularization can be defined as
\begin{equation*}
\mathcal{R}_4(\Theta) = \sum_{i}\sum_{j}\left( m^{t2s}_{ij} - \frac{1}{|S^s|}\right)^2,
\end{equation*}
where $m^{t2s}_{ij}$ is the probability of mapping the target slot $s^t_i$ to the source slot $s^s_j$ and $|S^s|$ denotes the number of source slots.

\section{Experiments}
\label{crossTL:Experiments}
In this section, we conduct experiments on simulation and real-world datasets to test the performance of the proposed PROMISE model.

\subsection{Baselines}
We compare the proposed PROMISE model with baseline methods, including
\begin{enumerate}
\item None-Transfer method (denoted by ``NoneTL''), which utilizes only the target domain dataset to train a dialogue policy that is based on the Gaussian process dialogue policy~\cite{gasic2014gaussian}.
\item Random Speech-act Mapping and Entropy slot-matching (denoted by ``RAFS''), which uses a random speech-act mapping and an entropy-based cross-domain slot matching~\cite{gavsic2015policy}.
\item Learned Speech-act Mapping and Entropy slot-matching (denoted by ``LAFS''), which uses a learned speech-act mapping and an entropy-based slot-matching matrix.
\item Perfect speech-act mapping and entropy-based slot-matching (denoted by ``FAFS''), which has the ground-truth speech-act mapping and the entropy-based slot-matching. This method is used to demonstrate the performance upper bound of the transfer learning.
\item Perfect speech-act mapping and learned slot-matching (denoted by ``FALS''), which has the ground-truth speech-act mapping, and the slot-matching is learned with the proposed algorithm. This method is used to demonstrate the upper bound of transfer learning when the slot mapping is known.
\end{enumerate}

\begin{figure*}[ht]
\centering\mbox{
\subfigure[Averaged Reward, higher is better.]{\label{crossTL:fig:sim_reward} \scalebox{0.66}{\includegraphics[width=\columnwidth]{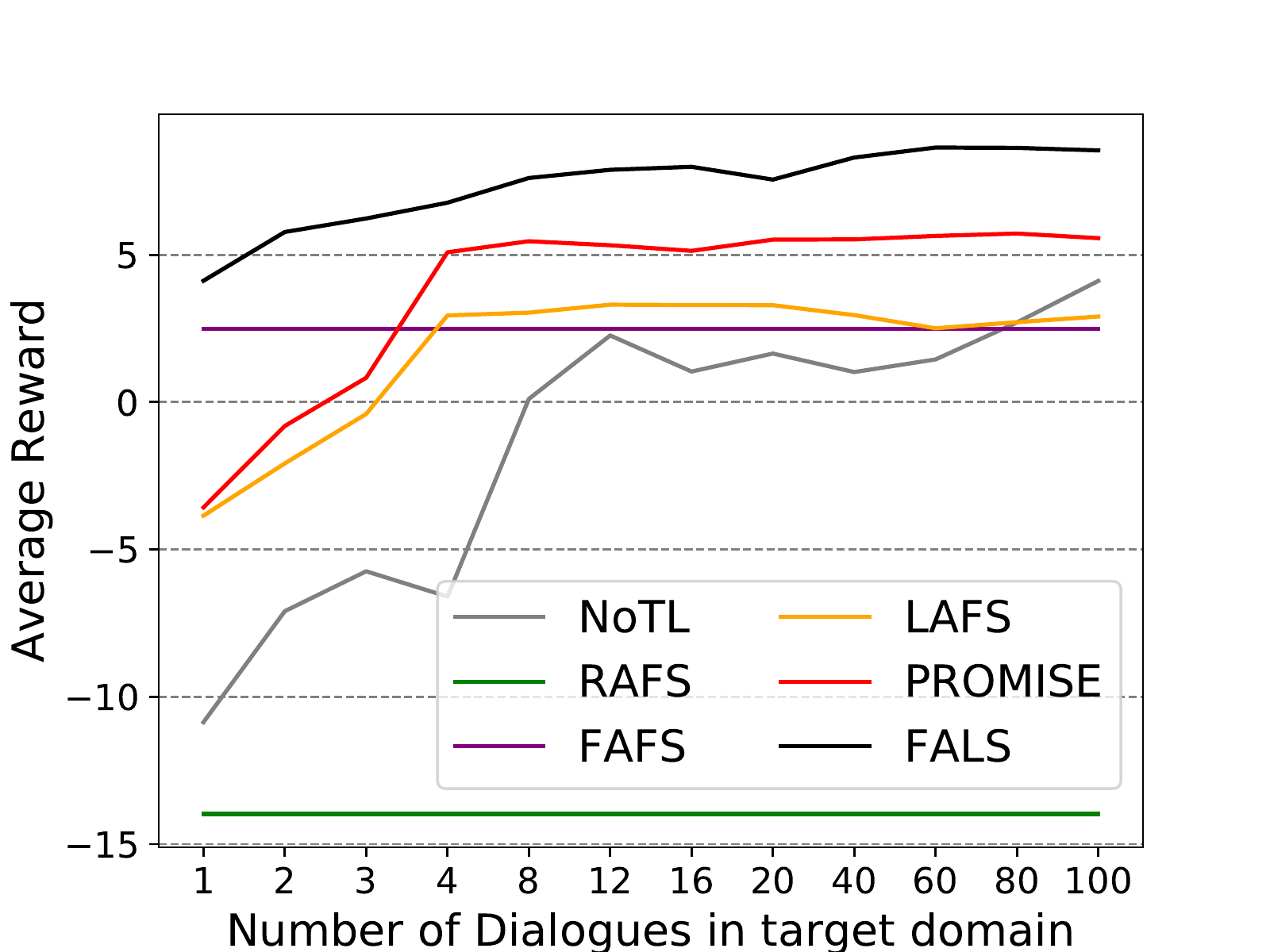}}}
\subfigure[Averaged Success Rate, higher is better.]{\label{crossTL:fig:sim_success} \scalebox{0.66}{\includegraphics[width=\columnwidth]{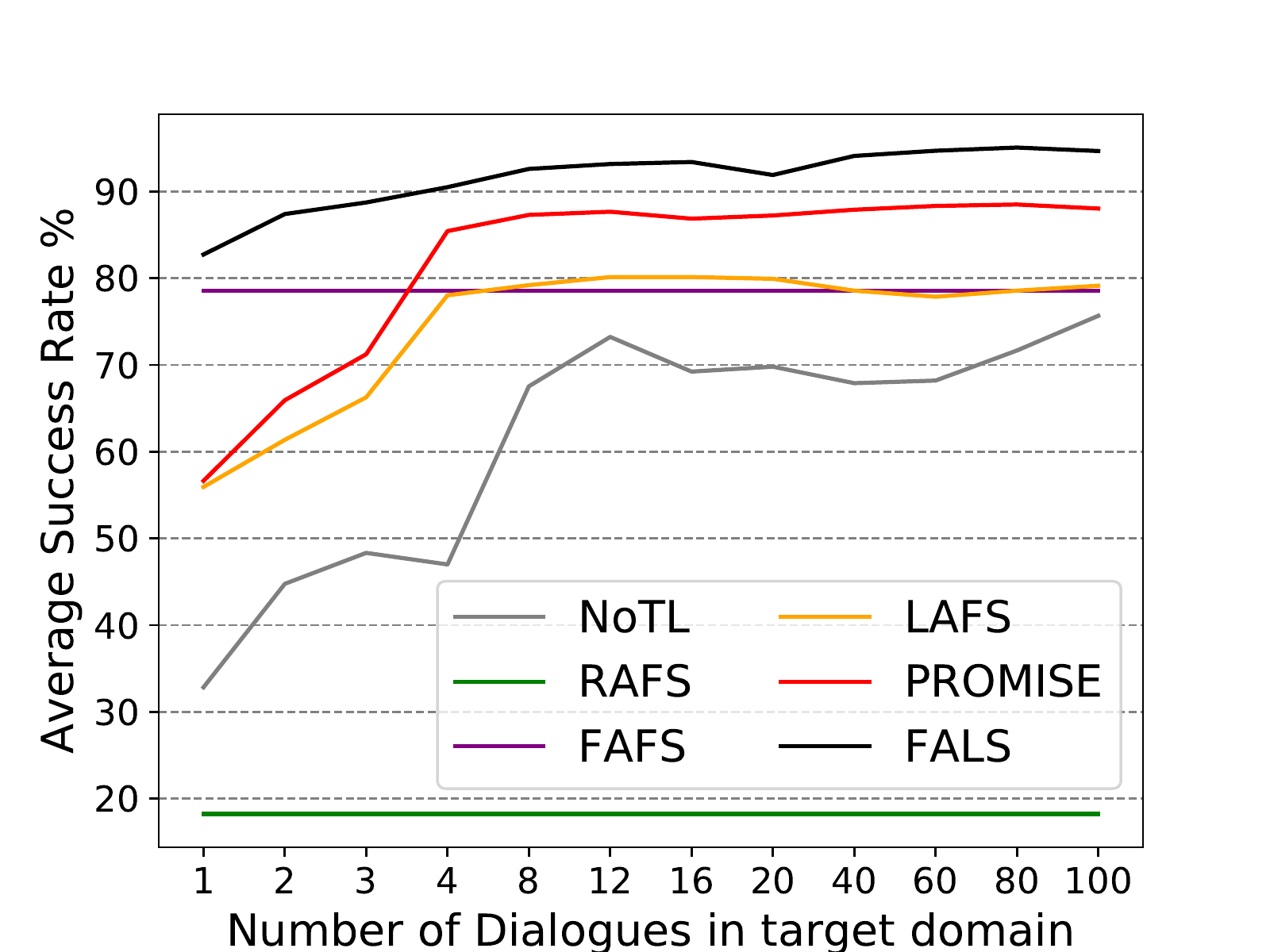}}}
\subfigure[Averaged Length, lower is better.]{\label{crossTL:fig:sim_length} \scalebox{0.66}{\includegraphics[width=\columnwidth]{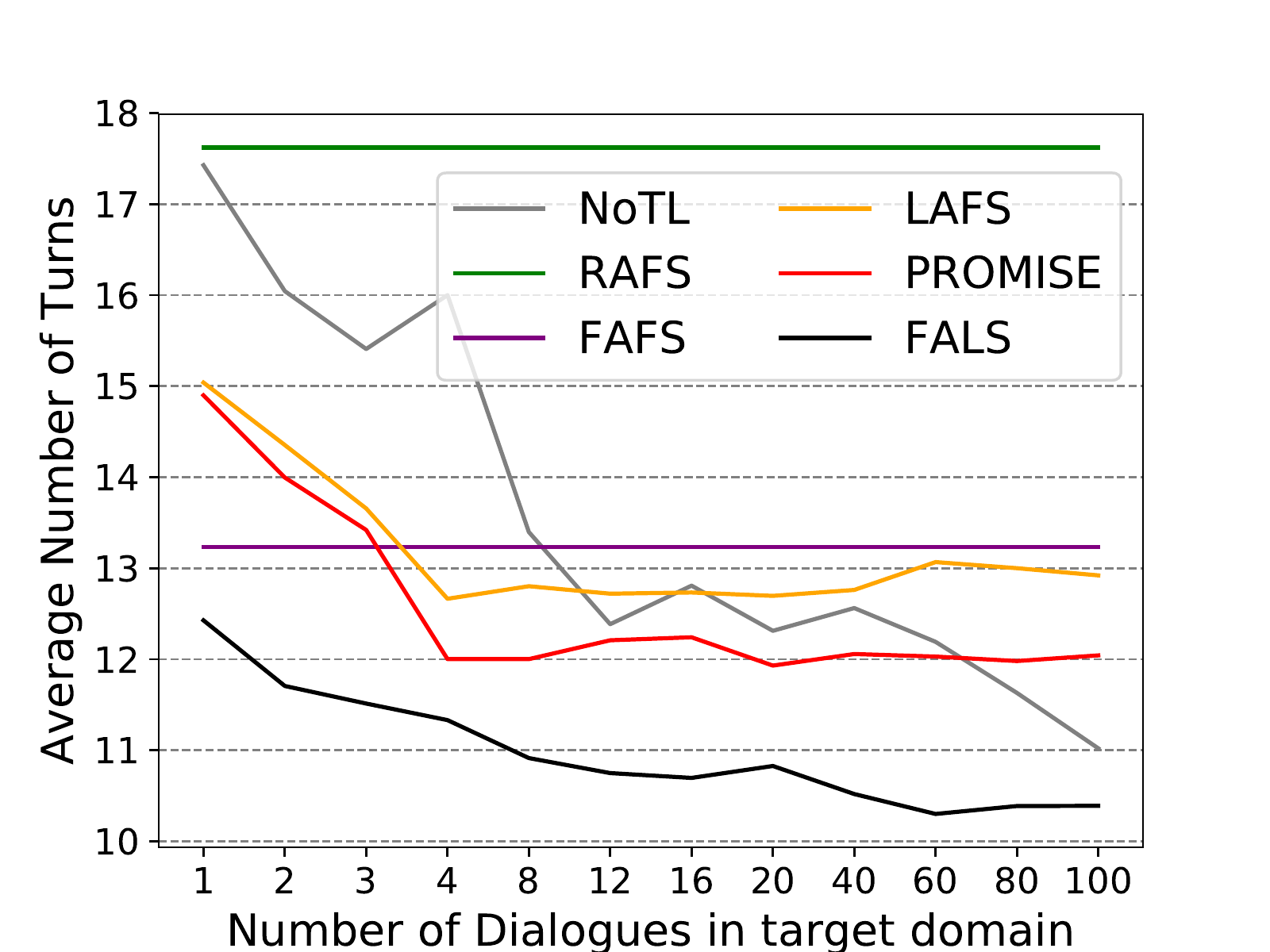}}}}
\caption{Experiment results for the simulation dataset. The methods ``FALS'' and ``FAFS'' are the performance upper-bounds since they use the ground truth speech-act mapping.}
\label{crossTL:fig:sim_reward_success}
\end{figure*}

\subsection{Source and Target Domains}

In the experiments, a dialogue policy in the Cambridge restaurants booking domain (denoted by ``CamRestaurants'') will be transferred to the target Cambridge hotel booking domain (denoted by ``CamHotels'').
The CamRestaurants domain is a restaurant booking dialogue system in Cambridge, and the CamHotels domain is a hotel booking dialogue system in Cambridge. The CamRestaurants domain has 9 slots including 4 informable slots which can be used to constraint the scope of search and 5 requestable slots which can only be requested after a venue is identified. For example, price is a informable slot since customer can ask for a cheap restaurant, but phone number is only requestable, because no customer will say ``I want to find a restaurant with phone number xxx''. The CamHotels domain has 6 informable slots and 5 requestable slots.

Note that the ground truth speech-act mapping and slot mapping are not available in our problem settings and so the algorithms do not know ``ack'' in the source domain corresponds to ``ack'' in the target domain, because the source and target domains could use different sets of speech-acts. Transferring dialogue policy across domains with different speech-acts requires learning the cross-domain speech-act mapping and slot mapping from the training data.

\begin{table}[h]
\caption{The number of dialogues used in each experiment is shown in the table. The simulation is repeated with $10$ different random seeds, and the real-world experiment is repeated for $5$ times with $5$ sets of target domain training data. Each set of real-world target domain training data have $20$ dialogues, so the total number of target domain training dialogues is $100$.}
\begin{center}
\tiny
\begin{tabular}{|c||c|c|c|}
  \hline
   & \multicolumn{2}{|c|}{Training} & \multicolumn{1}{|c|}{Testing} \\
  \hline
  Domain & \tabincell{c}{Source: \\CamRestaurants} & \tabincell{c}{Target: \\CamHotels} & \tabincell{c}{Target: \\CamHotels} \\
  \hline
  \tabincell{c}{Simulation \\(Total Reward)} & 1000 & 1-100 & Live: 300  \\
  \hline
  \tabincell{c}{Real \\(Immediate Reward)} & 40 & 1-20 & \tabincell{c}{Static: 60 \\Live: 300} \\
  \hline
\end{tabular}
\label{crossTL:tab:DataTab}
\end{center}
\end{table}

\begin{figure}[h]
\centering\mbox{
\subfigure[Speech-act mapping with only one target domain dialogue.]{\label{crossTL:fig:vis_actmap_1} \scalebox{0.5}{\includegraphics[width=\columnwidth]{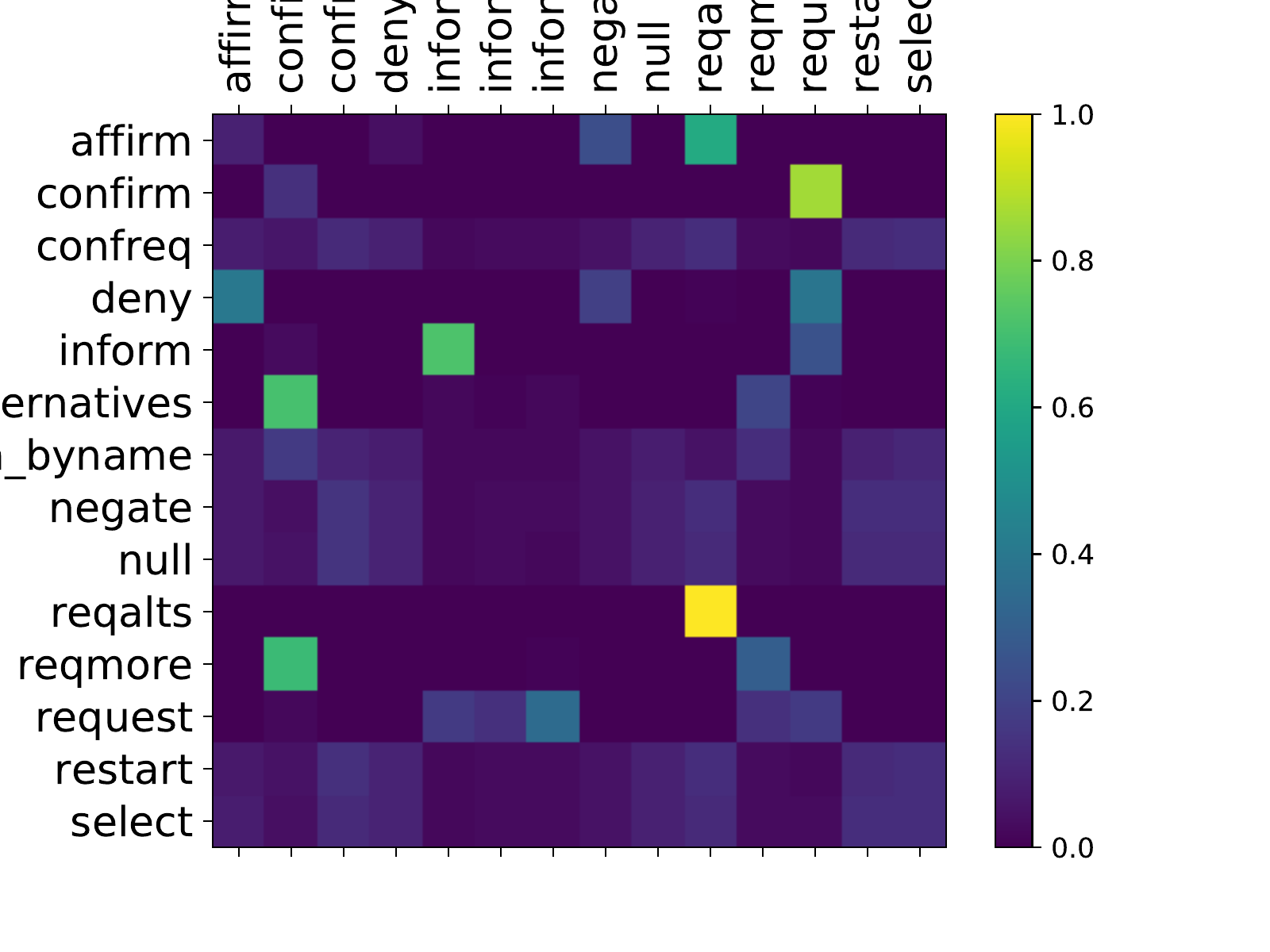}}}
\subfigure[Speech-act mapping with 100 target domain dialogue.]{\label{crossTL:fig:vis_actmap_100} \scalebox{0.5}{\includegraphics[width=\columnwidth]{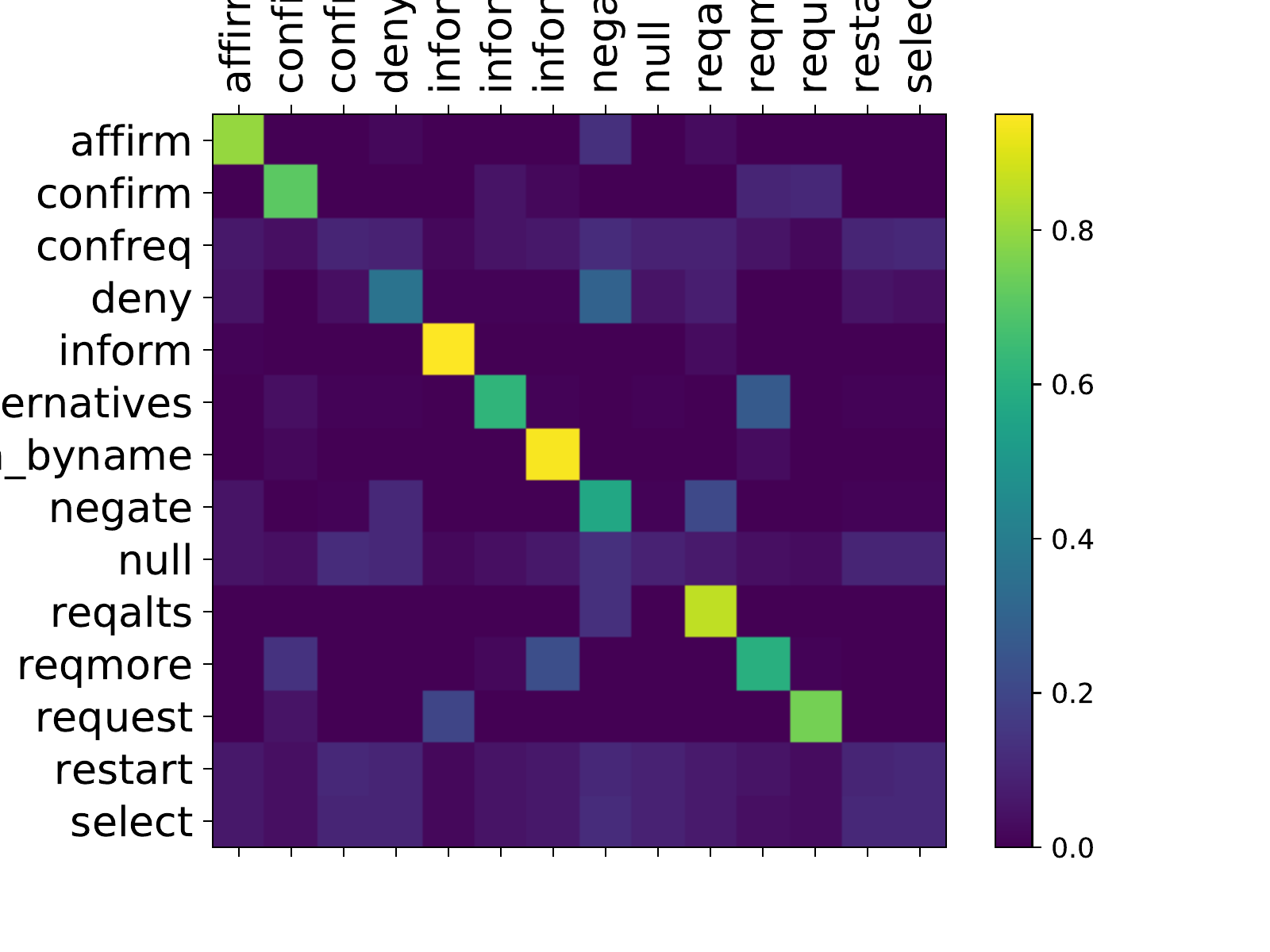}}}
}
\caption{Visualization for the speech-act mapping, where the ground truth speech-act mapping is the diagonal matrix.}
\label{crossTL:fig:vis_actmap}
\end{figure}

\subsection{Experiments on Simulation Data}

In this section, we conduct experiments on the user simulator in PyDial package~\cite{ultes2017pydial}.

\subsubsection{Experimental Settings}
In each dialogue, the simulator will randomly generate a set of requirements and it will try to ask for a venue that matches all requirements. 
In the simulation, only the final total reward will be used to train the dialogue policy and no immediate reward will be used. In this setting, less labelling effort is needed but more training dialogues are needed to train a good dialogue policy.
There are 1000 training dialogues in the source domain and we vary the number of training dialogues in the target domain from 1 to 100. The detailed statistics of the simulation dataset is listed in Table~\ref{crossTL:tab:DataTab}.

\subsubsection{Evaluation Metrics}
We use the averaged reward, the success rate, and the number of dialogue turns as performance metrics to evaluate the learned dialogue policies on $300$ test dialogues. We run the experiments with 10 different random seeds and report the averaged performance. The averaged reward is the mean of all rewards obtained by the simulated user while interacting with the agent being tested, the success rate measures the probability that the simulated user successfully orders a cup of coffee within $20$ turns, and the number of dialogue turns is the average number of dialogue turns before the simulated user can successfully complete an order.

\begin{figure*}[ht]
\centering\mbox{
\subfigure[Averaged AUC, higher is better.]{\label{crossTL:fig:real_auc} \scalebox{0.5}{\includegraphics[width=\columnwidth]{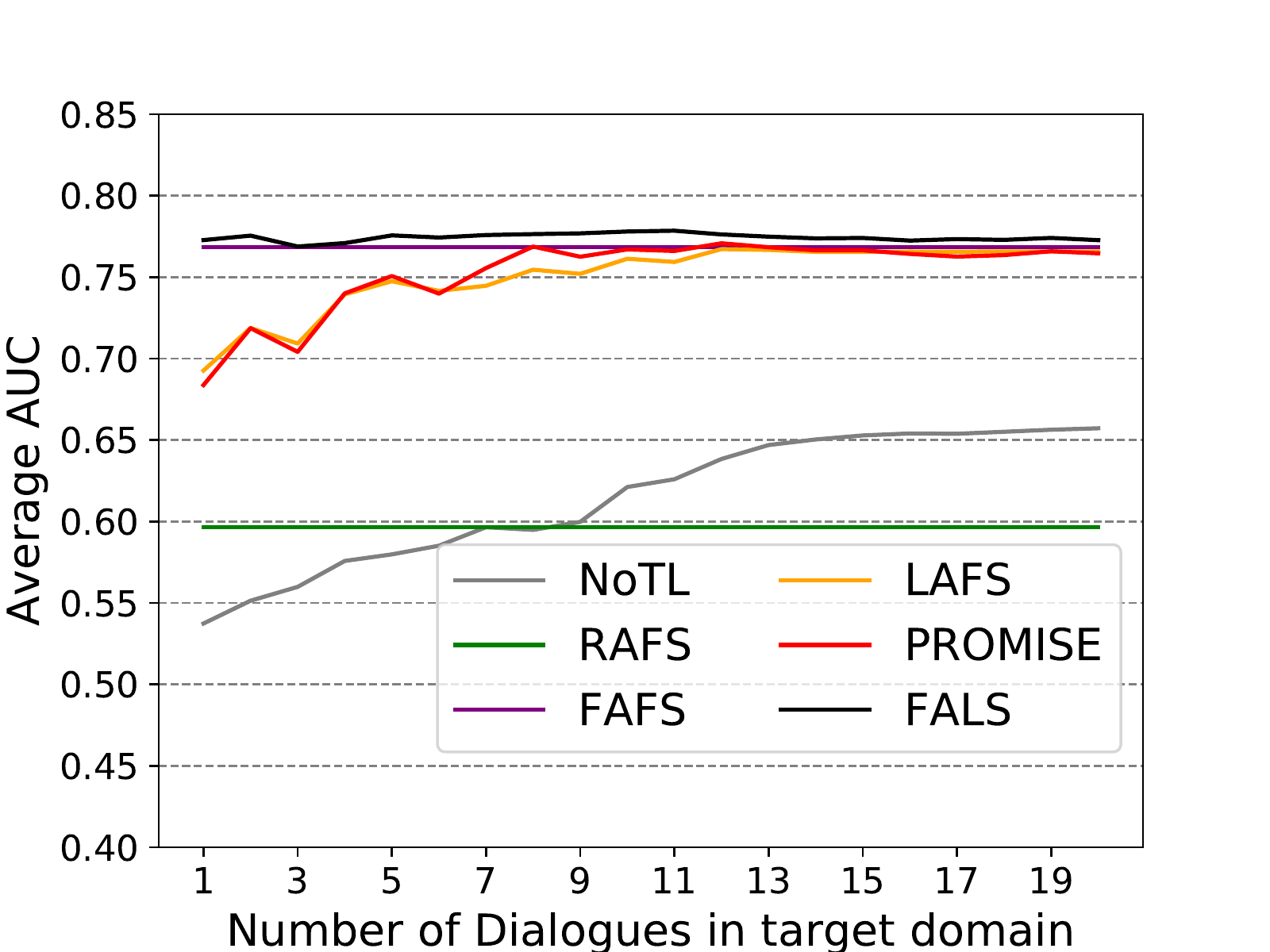}}}
\subfigure[Averagd Reward, higher is better.]{\label{crossTL:fig:real_reward} \scalebox{0.5}{\includegraphics[width=\columnwidth]{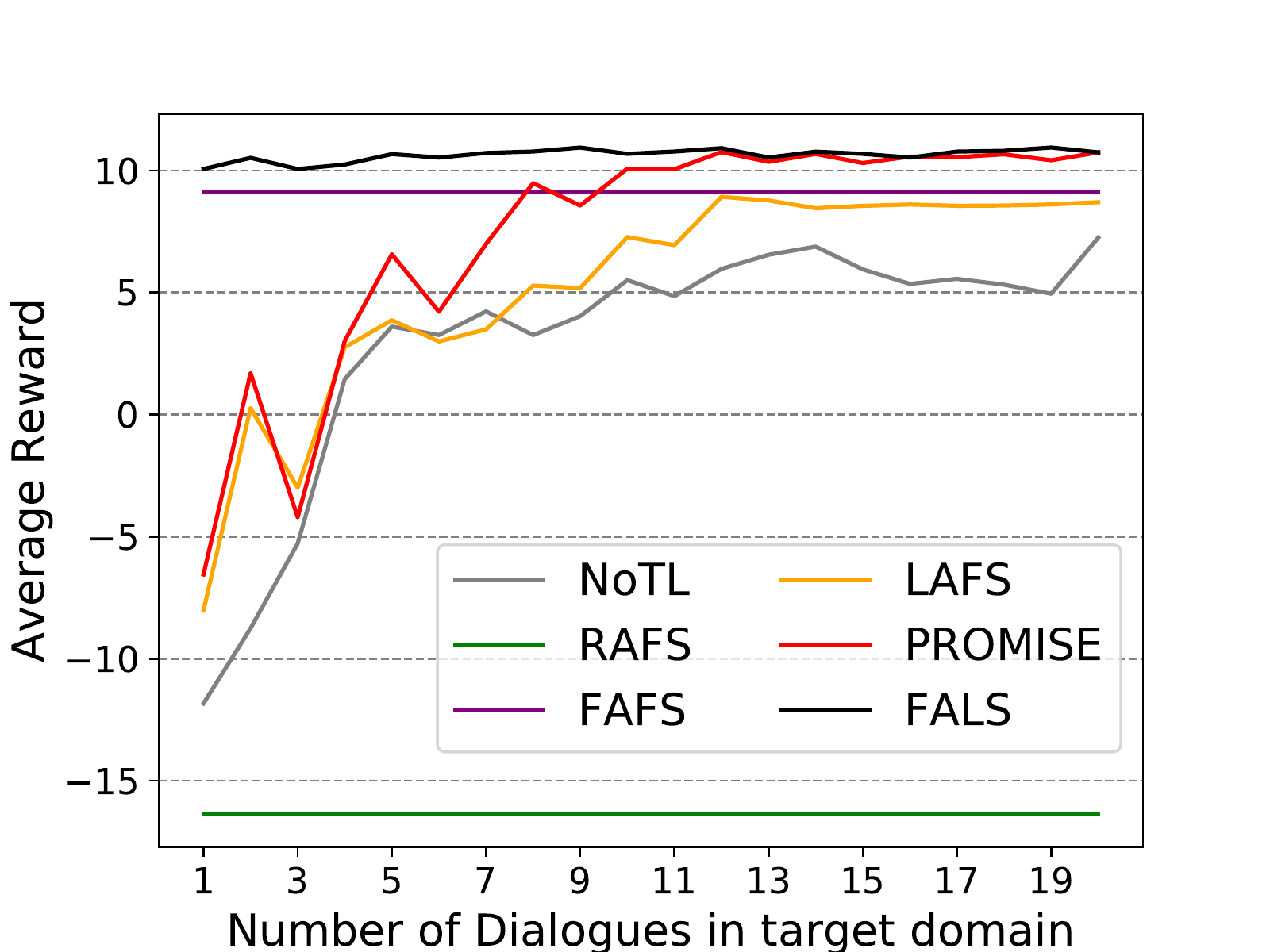}}}
\subfigure[Averaged Success Rate, higher is better.]{\label{crossTL:fig:real_success} \scalebox{0.5}{\includegraphics[width=\columnwidth]{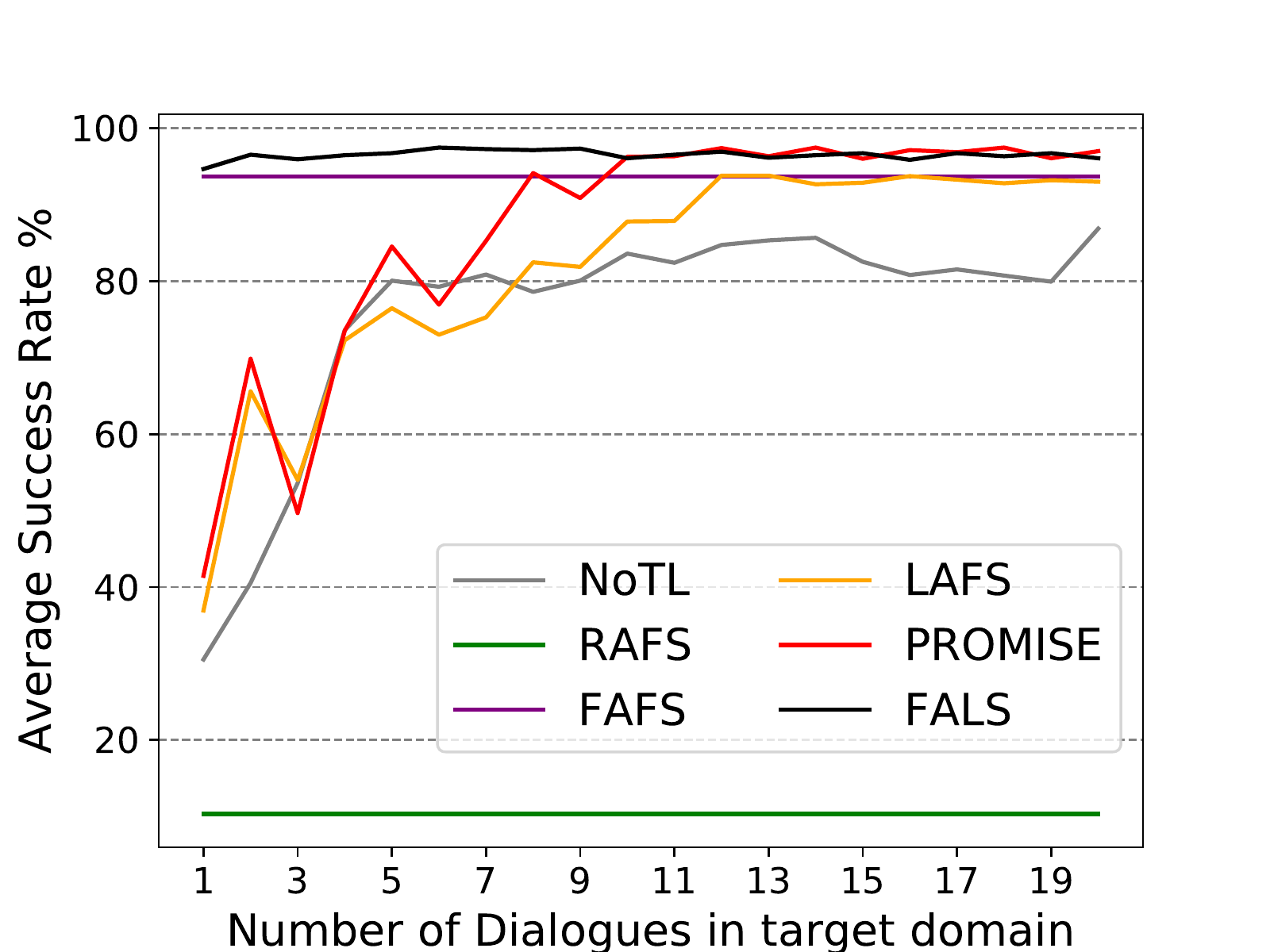}}}
\subfigure[Averaged Dialogue Length, lower is better.]{\label{crossTL:fig:real_length} \scalebox{0.5}{\includegraphics[width=\columnwidth]{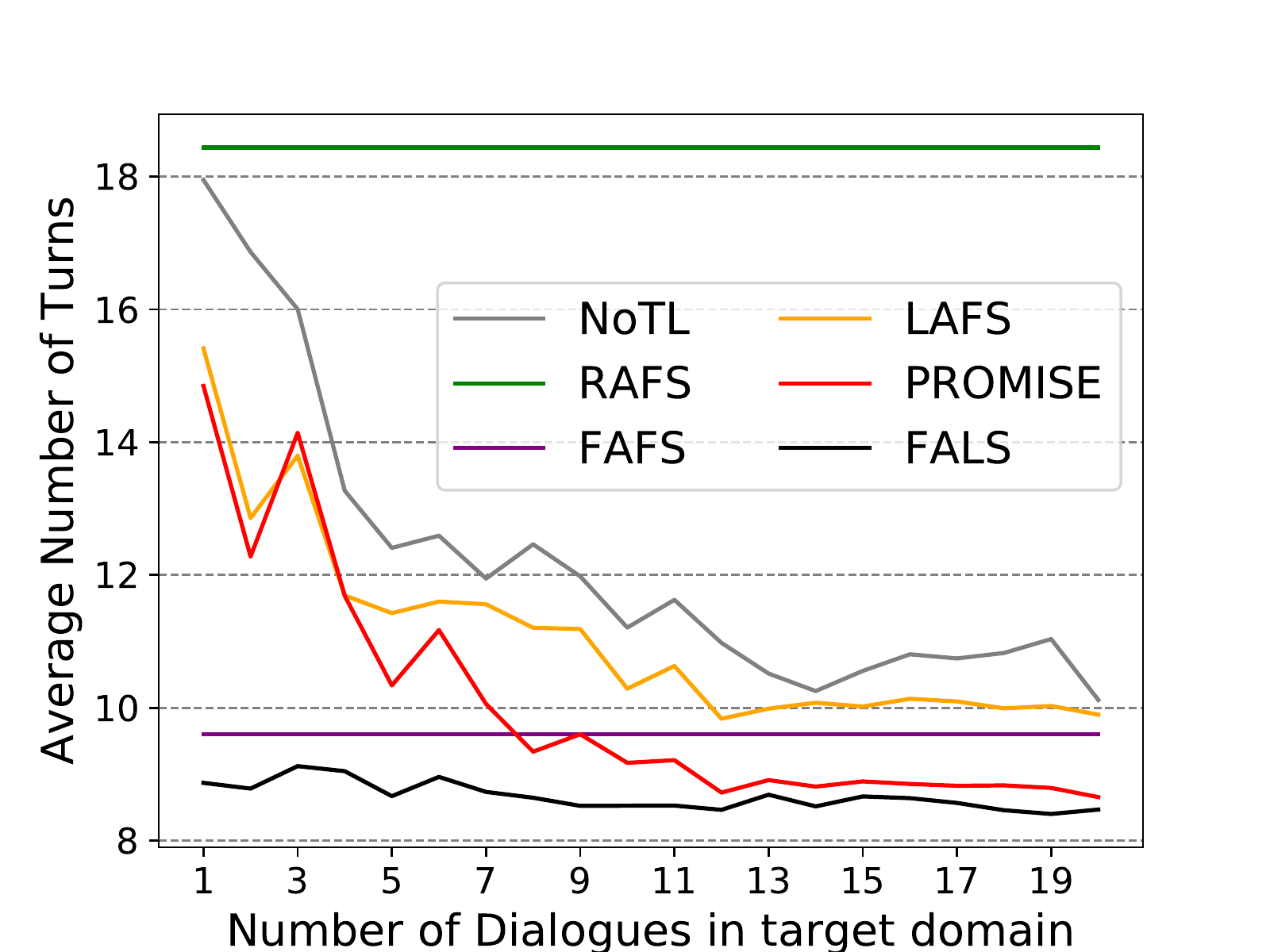}}}
}
\caption{Experiment results for the real-world dataset. The methods ``FALS'' and ``FAFS'' are the performance upper-bounds since they use the ground truth speech-act mapping. }
\label{crossTL:fig:real_auc_reward}
\end{figure*}

\subsubsection{Experimental Results}
The results are shown in Figure~\ref{crossTL:fig:sim_reward_success}. The performance of ``NoneTL'' increases slowly since it requires a large number of data to train a dialogue policy. The ``RAFS'' fails as it relies on a good speech-act mapping. As we can see, the proposed PROMISE model outperforms other baselines significantly, which demonstrates that learning speech-act and slot-mapping simultaneously is highly necessary and it can boost performance when there is not enough data in the target domain. PROMISE is better than ``LAFS'' which uses a heuristic slot-mapping and an external database, implying that learning the slot mapping is better than human \mbox{heuristics}.


\subsubsection{Visualization}

We visualize the learned cross-domain speech-acts mapping in Figure~\ref{crossTL:fig:vis_actmap} where the ground-truth speech-act mapping matrix should be a diagonal matrix since the speech-acts in two domains have one-to-one correspondence. As we can see, as more and more data are collected in the target domain, the learned speech-act mapping matrix is getting better.

\subsection{Experiments on Real-World Data}
In this section, we will conduct experiments on real-world data.

\subsubsection{Experimental Settings}
The training dataset is collected when human users are interacting with an online dialogue agent via the Wechat interface. After each dialogue, the human user gives a subjective feedback reward as a supervision signal. The dialogue agent is constantly improving its dialogue policy with a Gaussian process reinforcement learning model~\cite{gasic2014gaussian}. We have collected $40$ training dialogues in the source domain and $100$ dialogues in the target domain.
The test dataset is collected when two human users are interacting with each other, where one is acting the customer and the other is acting the dialogue agent. While one of the dialogue agent in the training dataset is a reinforcement learning algorithm, both the customer and the coffee servant are real persons in the test dataset. $60$ test dialogues are collected in the target domain as we only care about the performance in the target domain.
Five users participate in the data collection process, they also give immediate rewards on all dialogue turns and those immediate rewards will be used to train the dialogue policy. Compared with the simulation data, there are more rewards and hence fewer training dialogues are required to learn a satisfactory dialogue policy.
The detailed statistics of the real-world dataset is listed in Table~\ref{crossTL:tab:DataTab}.

\subsubsection{Evaluation Metrics}
We use two kinds of evaluation metrics to evaluate the dialogue policies trained on the real-world dataset.

A static evaluation is conducted on the test dataset because a good dialogue policy should have similar behaviours to the human agent. In each test dialogue turn, the dialogue agent ranks all candidate replies generated by the Pydial package, which is all the possible combinations of speech-acts and slots. The reply performed by the human annotator is labelled as 1 and the other candidate replies are labelled as 0. Then based on such labelling, the AUC score is used as a evaluation metric.
In each dialogue turn, we evaluate the Q-function for 10 times since the value of the Q-function is randomly sampled from a Gaussian distribution in the Gaussian process dialogue policy. We report the average AUC for all turns and for all samples.

A live evaluation is conducted with the user simulator in the Pydial package since a good dialogue policy should be able to serve the simulator user well. The user simulator in the Pydial package interacts lively with the dialogue agent for $300$ times where the averaged reward, the averaged success rate, and the averaged dialogue length are used as evaluation metrics.

\subsubsection{Experimental Results}
The results are shown in Figure~\ref{crossTL:fig:real_auc_reward}. In the static evaluation, we can see that the performance of ``NoTL'' increases slow. Moreover, we found that the ``NoTL'' has very large prediction variance due to the lack of training data.
The proposed PROMISE is much better than other baseline methods and very close to the performance upper-bound since it leverage source data.
These results show the proposed method can effective transfer dialogue policy across domains by learning speech-acts and slots mappings.

In the live evaluation, we can see that PROMISE significantly outperforms the ``NoTL'' method when there are more than 7 training dialogues in the target domain and its performance is close to the performance upper-bound when there are more than 12 training dialogues in the target domain. This result demonstrates the effectiveness of the proposed dialogue policy transfer method. Moreover, PROMISE outperforms the ``LAFS'' method, which again demonstrates the effectiveness of learning the slot mapping.

\section{Conclusion}
In this paper, we propose the PROMISE model to tackle the problem of transferring dialogue policies across domains with different speech-acts and disjoint slots.
The PROMISE model can learn to align different speech-acts and slots simultaneously, and it does not require common slots or additional database to calculate slot entropies.
Experiments demonstrate that PROMISE can effectively transfer dialogue policies. In the future, we plan to investigate how to transfer from multiple source domains simultaneously.

\bibliographystyle{named}
\bibliography{PROMISE}

\end{document}